\documentclass{article}
\usepackage{spconf,amsmath,graphicx}

\usepackage{cite} 
\usepackage{microtype} 

\usepackage{tabularx}
\usepackage{booktabs}
\usepackage{multirow}

\newcolumntype{C}{>{\centering\arraybackslash}X}
\newcolumntype{L}{>{\raggedright\arraybackslash}X}
\newcolumntype{b}{X}
\newcolumntype{s}{>{\hsize=.2\hsize}X}
\newcolumntype{e}{>{\hsize=.4\hsize}X}
\newcolumntype{f}{>{\hsize=.5\hsize}X}

\newcommand{\textunderscript}[1]{$_{\text{#1}}$}

\usepackage{mathtools}

\usepackage{xcolor}
\usepackage{url}
\usepackage[skip=2pt]{caption}  
\usepackage{tikz}

\newcommand{\rotcol}[1]{\begin{tabular}{@{}c@{}}
\hspace{-1mm}{\rotatebox[origin=c]{90}{#1}}\hspace{-1mm}
\end{tabular}}

\title{Analyzing ASR pretraining for low-resource speech-to-text translation}
\name{Mihaela C. Stoian, Sameer Bansal, Sharon Goldwater}
\address{
  School of Informatics, University of Edinburgh, UK \\
{\small \tt \{c.mihaela.stoian, sameer.bansal\}@ed.ac.uk, sgwater@inf.ed.ac.uk} \\ 
\\}
\begin{document}
\ninept 
\maketitle
\begin{abstract}
 
Previous work has shown that for low-resource source languages, automatic speech-to-text translation (AST) can be improved by pretraining an end-to-end model on automatic speech recognition (ASR) data from a high-resource language. However, it is not clear what factors---e.g., language relatedness or size of the pretraining data---yield the biggest improvements, or whether pretraining can be effectively combined with other methods such as data augmentation. Here, we experiment with pretraining on datasets of varying sizes, including languages related and unrelated to the AST source language. We find that the best predictor of final AST performance is the word error rate of the pretrained ASR model, and that differences in ASR/AST performance correlate with how phonetic information is encoded in the later RNN layers of our model. We also show that pretraining and data augmentation yield complementary benefits for AST.

\end{abstract}
\section{Introduction}
\label{sec:intro}

Low-resource automatic speech-to-text translation (AST) has recently 
gained traction
as a way to bring NLP tools to under-represented languages. An end-to-end approach \cite{berard+etal_nipsworkshop16, weiss2017sequence, bansal2018interspeech, berard2018end, bansal+pretraining+arxiv+2018, sperber19tacl, salesky-etal-2019-fluent} is particularly appealing for source languages with no written form, or for endangered languages where translations into a high-resource language may be easier to collect than transcriptions \cite{mboshi+french+corpora}.
However, building high-quality end-to-end AST with little parallel data is challenging, and has led researchers to explore 
how other sources of data could be used to help.

A number of methods have been investigated. Several of these use transcribed source language audio and/or translated source language text in a multitask learning scenario~\cite{antonis+tied+naacl18, berard2018end, sperber19tacl}
or to pre-train parts of the model before fine-tuning on the end-to-end AST task \cite{berard2018end}.
Others assume, as we do here, that no additional source language resources are available, in which case transfer learning using data from language(s) other than the source language is a good option.
In particular, several researchers have shown that low-resource AST can be improved by pretraining on an ASR task in some other language, then transferring the encoder parameters to initialize the AST model.  For example, Bansal et al. \cite{bansal+pretraining+arxiv+2018} showed that pre-training on either English or French ASR improved their Spanish-English AST system (trained on 20 hours of parallel data) and Tian \cite{yusheng2019_msc} 
got improvements on an 8-hour Swahili-English AST dataset using English ASR pretraining.

Overall these results show that pretraining helps, but leave 
open the question of what factors affect the degree of improvement. For example, does language relatedness play a role, or simply the amount of pretraining data?  Bansal et al. showed bigger AST gains as the amount of English pretraining data increased from 20 to 300 hours, and also found a slightly larger improvement when pretraining on 20 hours of English versus 20 hours of French, but they pointed out that the Spanish data contains many English code-switched words, which could explain the latter result. In related work on multilingual pretraining for low-resource ASR,  Adams et al. \cite{adams-etal-2019-massively} showed that pre-training on more languages helps, but it is not clear whether the improvement is due to including more languages, or just more data.

To begin to tease apart these issues, we focus here on monolingual pretraining for low-resource AST, and investigate two questions. First, can we predict what sort of pretraining data is best for a particular AST task? Does it matter if the pretraining language is related to the AST source language (defined here as part of the same language family, since phonetic similarity is difficult to measure), or is the amount of pretraining data (or some other factor) more important? Second, can pretraining be effectively combined with other methods, such as data augmentation, in order to further improve AST results?

To answer these questions, we use the same AST architecture and Spanish-English parallel data as Bansal et al.~\cite{bansal+pretraining+arxiv+2018}, but pretrain the encoder using a number of different ASR datasets: the 150-hour AISHELL corpus of Chinese as well as seven GlobalPhone languages, each with about 20 hours of data. We find that pretraining on a larger amount of data from an unrelated language is much better than pretraining on a smaller amount of data from a related language. Moreover, even when controlling for the amount of data, the WER of the ASR model from pretraining seems to be a better predictor of final AST performance than does language relatedness. Indeed, we show that there is a very strong correlation between the WER of the pretraining model and BLEU score of the final AST model---i.e., the best pretraining strategy may simply be to use datasets and methods that will yield the lowest ASR WER during pretraining. However, we also found that AST results can be improved further by augmenting the AST data using standard speed perturbation techniques \cite{povey-3-way}. Our best results using non-English pretraining data improve the test set BLEU scores of an AST system trained on 20 hours of parallel data from 10.2 to 14.3, increasing to 15.8 with data augmentation. 

Finally, we analyze the representations learned by the models and show that better performance seems to correlate with 
the extent to which  phonetic information is encoded in a linearly separable way in the later RNN layers.

\section{Methodology}
\label{sec:methodology}

\begin{figure}[t]  
\begin{minipage}[t]{1.0\linewidth}
    \centering
     \centerline{\includegraphics[scale=0.6]{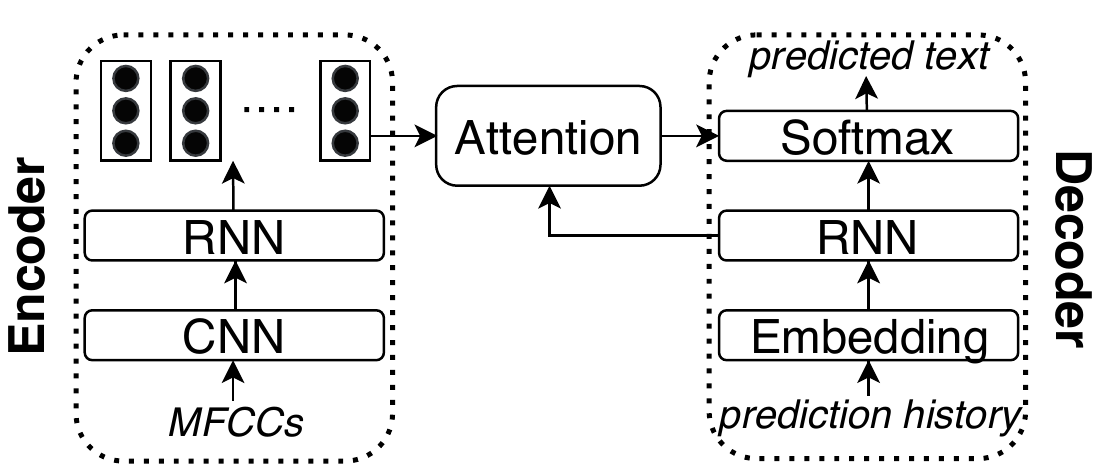}}
    \caption{Encoder-decoder architecture used for both ASR and AST. 
    }
    \label{diagram:enc-dec}
\end{minipage}
\end{figure}

For both ASR and AST tasks we use the same end-to-end system architecture shown in Figure \ref{diagram:enc-dec}: the encoder-decoder model from \cite{bansal+pretraining+arxiv+2018}, which itself is adapted from \cite{weiss2017sequence}, \cite{berard2018end} and \cite{bansal2018interspeech}. Details of the architecture and training parameters are described in Section~\ref{sec:training}.

After pretraining an ASR model, we transfer only its encoder parameters to the AST task.
Previous experiments \cite{bansal+pretraining+arxiv+2018} showed that the encoder accounts for most of the benefits of transferring the parameters. Transferring also the decoder and attention mechanism does bring some improvements, but is only feasible when the ASR pretraining language is the same as the AST target language, which is not true in most of our experiments. 

In addition to pretraining, we experimented with data augmentation. 
Specifically, we augmented the AST data using Kaldi's \cite{povey+kaldi+asru+2011} 3-way speed perturbation,
adding versions of the AST data 
where the audio is sped down and up by a factor of 0.9 and 1.1, respectively.\footnote{In principle, we can augment the ASR pretraining data, the AST data, or both. However, we only  augmented the AST data because in a preliminary experiment on AISHELL, we found that
augmenting the ASR pretraining data did not improve its WER or the performance of the final AST system. Other researchers have reported ASR improvements using speed perturbation, and given the strong correlation we report below between ASR WER and AST BLEU, we would expect other data augmentation methods that do improve WER in pre-training to also improve AST.} 

To evaluate ASR performance we compute the word error rate (WER).\footnote{\url{https://github.com/belambert/asr-evaluation}} To evaluate AST performance we calculate the 4-gram BLEU score \cite{papineni+bleu} on four reference translations.\footnote{\url{https://www.nltk.org/_modules/nltk/translate/bleu_score.html}}

\section{Experimental Setup}

\subsection{Parallel data}
For the AST models, we use Spanish-English parallel data from Fisher corpus~\cite{post2014fisher}, containing 160 hours of Spanish telephone speech translated into English text. To simulate low-resource settings, we randomly downsample the original corpus to 20 hours of training data. Each of the dev and test sets comprise 4.5 hours of speech.

\subsection{Pretraining data}
Since we focus on investigating factors that might affect the AST improvements over the baseline when pretraining, we have chosen ASR datasets for pretraining that contrast in the number of hours and/or in the language similarity with Spanish.
Statistics for each dataset are in the left half of Table~\ref{tab:data-results}, with further details below.

To look at a range of languages with similar amounts of data, we used {\bf GlobalPhone corpora from seven languages} \cite{schultz2002globalphone}, each with around 20 hours of speech: Mandarin Chinese (zh), Croatian (hr), Czech (cs), French (fr), Polish (pl), Portuguese (pt), and Swedish (sv). French and Portuguese, like the source language (Spanish), belong to the Romance family of languages, while the other languages are less related---especially Chinese, which is not an Indo-European language. GlobalPhone consists of read speech recorded using similar conditions across languages, and the transcriptions for Chinese are Romanized, with annotated word boundaries.

To explore the effects of using a large amount of pretraining data from an unrelated language, we used the  
{\bf AISHELL-1 corpus of Mandarin Chinese}~\cite{bu2017aishell}, which contains 150 hours of read speech. Transcriptions with annotated word boundaries are available in both Hanzi (Chinese characters) and Romanized versions, and we built models with each. To compare to the GlobalPhone data, we also created a 20-hour subset of the Romanized AISHELL (\emph{zh-ai-small}) by randomly selecting utterances from a subset of the speakers (81, roughly the number present in most of the GlobalPhone datasets).

Finally, to reproduce one of the experiments from \cite{bansal+pretraining+arxiv+2018}, we pre-trained one model using 300 hours of {\bf Switchboard English} \cite{LDC97S62}. This data is the most similar to the AST speech data in terms of style and channel (both are conversational telephone speech). However, as noted by \cite{bansal+pretraining+arxiv+2018}, the Fisher Spanish speech contains many words that are actually in English (code-switching), so pretraining on English may provide an unfair advantage relative to other languages.

\begin{table}[t]
  \begin{center} \small
  \begin{tabularx}{0.95\linewidth}{rrr|cl}
    \toprule
     \multicolumn{3}{r}{{\bf DATA}} \vrule & \multicolumn{2}{l}{\bf RESULTS} \\ \midrule 
      {\bf Dataset}  & {\bf Hrs.}  & {\bf Spks.} & \bf ASR (WER) & \bf AST (BLEU)\\
           \midrule
           ast-20h & 20 &  & | & 10.3 \\
           \midrule
    zh-ai-small & 20 & 81 & 38.7  & 12.4 (+2.1)\\
    zh-ai-large  & 150 &  340 & 22.5 & 14.6 (+4.3)\\
    zh-ai-hanzi  & 150 &  340 &  25.3 & 13.2 (+2.9)\\
    \midrule
    hr-gp    & 12 & 72& 71.5 & 10.7 (+0.4)\\
    sv-gp  & 18 & 79 & 59.4 & 12.3 (+2.0)\\
    pl-gp & 19 & 79 & 59.6 & 10.8 (+0.5)\\
    pt-gp & 23 & 86 & 80.5 & 10.5 (+0.2)\\
    fr-gp  & 25 & 84 & 31.1 & 12.5 (+2.2)\\
    zh-gp  & 26 & 111 &  51.5 & 12.0 (+1.7)\\
    cs-gp    & 27 & 82& 53.7 & 11.1 (+0.8)\\
    \midrule 
    multilin6 & 124 & 482 & 44.2 & 13.3 (+3.0)\\ 
    \bottomrule 
  \end{tabularx}
  \end{center}
  \caption{Dataset statistics (left); dev set results from ASR pretraining and from the final AST system (right).
  AST results in all rows except the first are from pretraining using the dataset listed in that row, followed by fine-tuning using \emph{ast-20h}.
  Numbers in brackets are the improvement over the baseline.}
  \label{tab:data-results}
\end{table}

\subsection{Preprocessing}
We compute 13-dim MFCCs and cepstral mean and variance normalization along speakers using Kaldi \cite{povey+kaldi+asru+2011} on our ASR and AST audio. To shorten the training time, we trimmed utterances from the AST data to 16 seconds (or 12 seconds for the 160h augmented dataset). \looseness=-1

To account for unseen words in the test data, we model the ASR and AST text outputs via sub-word units using byte-pair encoding (BPE)~\cite{sennrich-haddow-birch:2016:P16-12}. We do this separately for each dataset as BPE works best as a language-specific tool (i.e. it depends on the frequency of different subword units, which varies with the language). We use 1k merge operations in all cases except Hanzi, where there are around 3000 symbols initially (vs around 60 in the other datasets). For Hanzi we ran experiments with both 1k and 15k merge operations. For Chinese Romanized transcriptions we removed tone diacritics.

\subsection{Model architecture and training} \label{sec:training} 
Following the architecture and training procedure described in~\cite{bansal+pretraining+arxiv+2018}, input speech features are fed into a stack of two CNN layers. In each CNN layer we stride the input with a factor of 2 along time, apply ReLU activation \cite{nair2010rectified} followed by batch normalization~\cite{ioffe+batchnorm+arxiv_2015}. The CNN output is fed into a three-layer bi-directional long short-term memory network (LSTM)~\cite{hochreiter+lstm}, with 512 hidden layer dimensions. 
For decoding,  
we use the predicted token 20\% of the time and the training token 80\% of the time~\cite{williams+teacher_forcing} as input to a 128-dimensional embedding layer followed by a three-layer LSTM, with 256 hidden layer dimensions, and combine this with the output from the attention mechanism~\cite{luong2015effective} to predict the word at the current time step.

We use code and hyperparameter settings from~\cite{bansal+pretraining+arxiv+2018}\footnote{ \url{https://github.com/0xSameer/ast}.}: the Adam optimizer~\cite{kingma+adam+arxiv+2014} with an initial learning rate of 0.001 and decay it by a factor of 0.5 based on the dev set BLEU score.
When training AST models, we regularize using dropout~\cite{srivastava+dropout} with a ratio of $0.3$ over the embedding and LSTM layers~\cite{Gal2015Theoretically}; weight decay with a rate of $0.0001$; and, after the first 20 epochs, 30\% of the time we replace the predicted output word by a random word from the target vocabulary.
At test time we use beam decoding with a beam size of 5
and length normalization~\cite{wu2016google+length+norm} with a weight of 0.6.

\section{Results and Discussion}

\begin{figure}[t]
  \centering
  \includegraphics[width=0.8\linewidth]{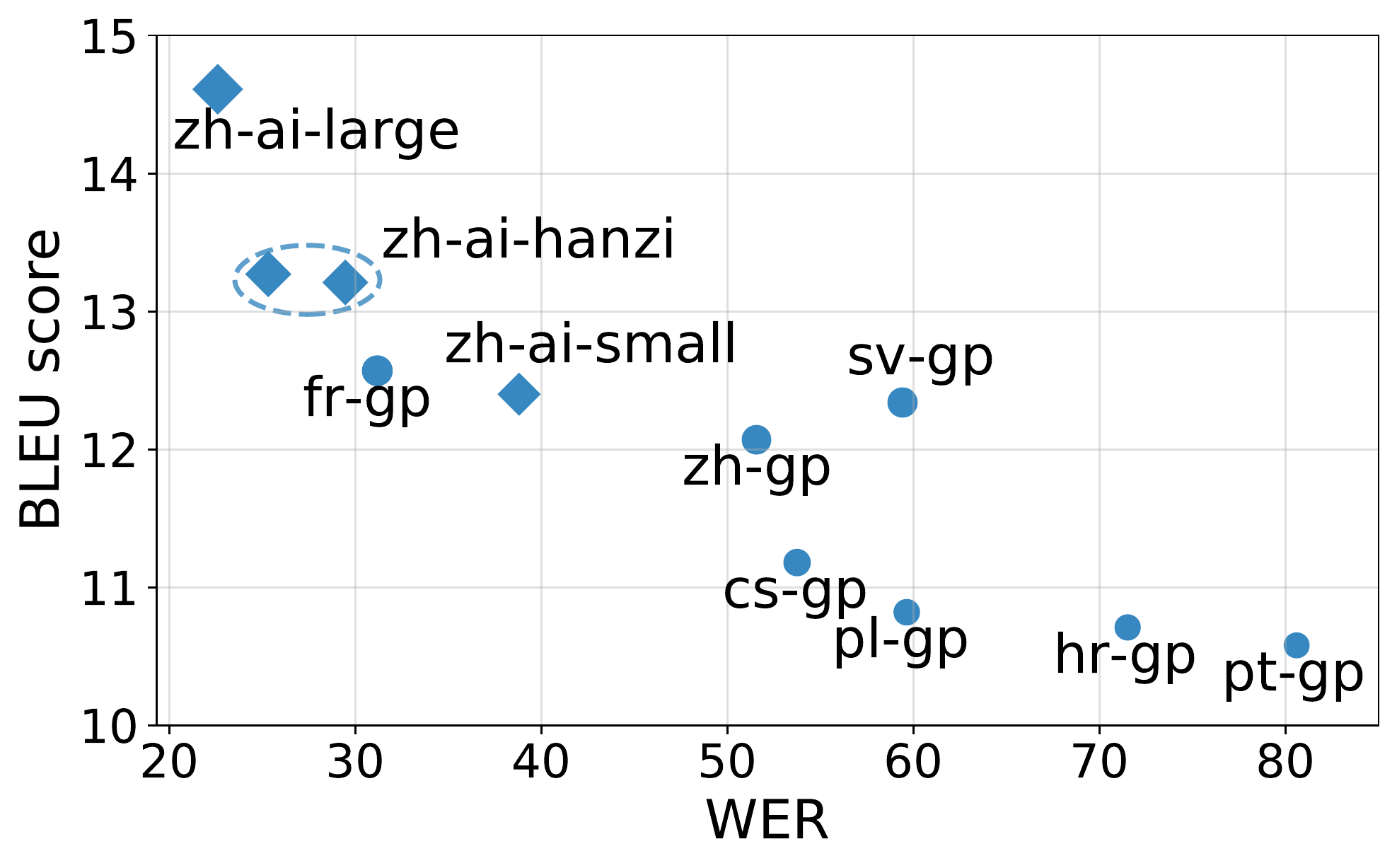}
  \caption{WER of each ASR model vs BLEU score of the corresponding pre-trained AST model, computed in both cases on dev sets. Diamond markers are AISHELL data sets; circles are from GlobalPhone. The points in the circled group come from different runs on the same dataset but with different BPE or learning rate schedules. The Spearman rank correlation of these points is -0.97; the correlation is -0.92 when using test sets to compute both ASR and BLEU.}
  \label{fig:asr-ast-strong-corr-mono}
\end{figure}

\subsection{Baseline and ASR results}

Our baseline 20-hour AST system obtains a BLEU score of 10.3 (Table~\ref{tab:data-results}, first row), 0.5 BLEU point lower than that reported by \cite{bansal+pretraining+arxiv+2018}. This discrepancy might be due to differences in subsampling from the 160-hour AST dataset to create the 20-hour subset, or from Kaldi parameters when computing the MFCCs.

WERs for our pre-trained models (Table~\ref{tab:data-results}) vary from 22.5 for the large AISHELL dataset with Romanized transcript to 80.5 for Portuguese GlobalPhone. These are considerably worse than state-of-the-art ASR systems (e.g., Kaldi recipes can achieve WER of 7.5 on AISHELL and 26.5 on Portuguese GlobalPhone), but we did not optimize our architecture or hyperparameters for the ASR task since our main goal is to analyze the relationship between pretraining and AST performance (and in order to use pretraining, we must use a seq2seq model with the architecture as for AST).

\subsection{Pretraining the AST task on ASR models}

AST results for our pre-trained models are given in Table~\ref{tab:data-results}. Pretraining improves AST performance in every case, with improvements ranging from 0.2 (\emph{pt-gp}) to 4.3 (\emph{zh-ai-large}). These results make it clear that language relatedness does not play a strong role in predicting AST improvements, since on the similar-sized GlobalPhone datasets, the two languages most related to Spanish (French and Portuguese) yield the highest and lowest improvements, respectively. Moreover, pretraining on the large Chinese dataset yields a bigger improvement than either of these---4.3 BLEU points. This is nearly as much as the 6 point improvement reported by \cite{bansal+pretraining+arxiv+2018} when pretraining on 100 hours of English data, which is especially surprising given not only that Chinese is very different from Spanish, but also that the Spanish data contains some English words.

This finding seems to suggest that data size is more important than language relatedness for predicting the effects of pretraining. However, there are big differences even amongst the languages with similar amounts of pretraining data. Analyzing our results further, we found a striking correlation between the WER of the initial ASR model and the BLEU score of the AST system pretrained using that model, as shown in Figure \ref{fig:asr-ast-strong-corr-mono}.
Therefore, although pretraining data size clearly influences AST performance, this appears to be mainly due to its effect on WER of the ASR model. We therefore hypothesize that WER is a better direct predictor of AST performance than either data size or language relatedness.

\subsection{Multilingual pretraining}

Although our main focus is monolingual pretraining, we also looked briefly at multilingual pretraining, inspired by
recent work on multilingual ASR 
\cite{toshniwal2018multilingual, zhou2018multilingual} and evidence that multilingual pretraining followed by fine-tuning on a distinct target language can improve ASR on the target language \cite{adams-etal-2019-massively, cho2018multilingual, dalmia2018sequence}. These experiments did not directly compare pretraining using a similar amount of monolingual data, but such a comparison was done by \cite{hermann2018multilingual, enno2018multilingual} in their work on learning feature representations for a target language with no transcribed data. They found a benefit for multilingual vs monolingual pretraining given the same amount of data.

Following up on this work, we tried pretraining using 124 hours of multilingual data (all GlobalPhone languages except Chinese), roughly the amount of data in our large Chinese models. We combined all the data together and trained an ASR model using a common target BPE with 6k merge operations, then transferred only the encoder to the AST model. However, we did not see a benefit to the multilingual training (Table~\ref{tab:data-results}, final row); in fact the resulting AST model was slightly worse than the \emph{zh-ai-large} model (BLEU of 13.3 vs 14.6). Other configurations of multilingual training might still outperform their monolingual counterparts, but we leave this investigation as future work.

\subsection{Augmenting the parallel data} 
 
\begin{figure}[t]
 \begin{minipage}[t]{1.0\linewidth}
    \centering
     \centerline{\includegraphics[width=0.8\linewidth]{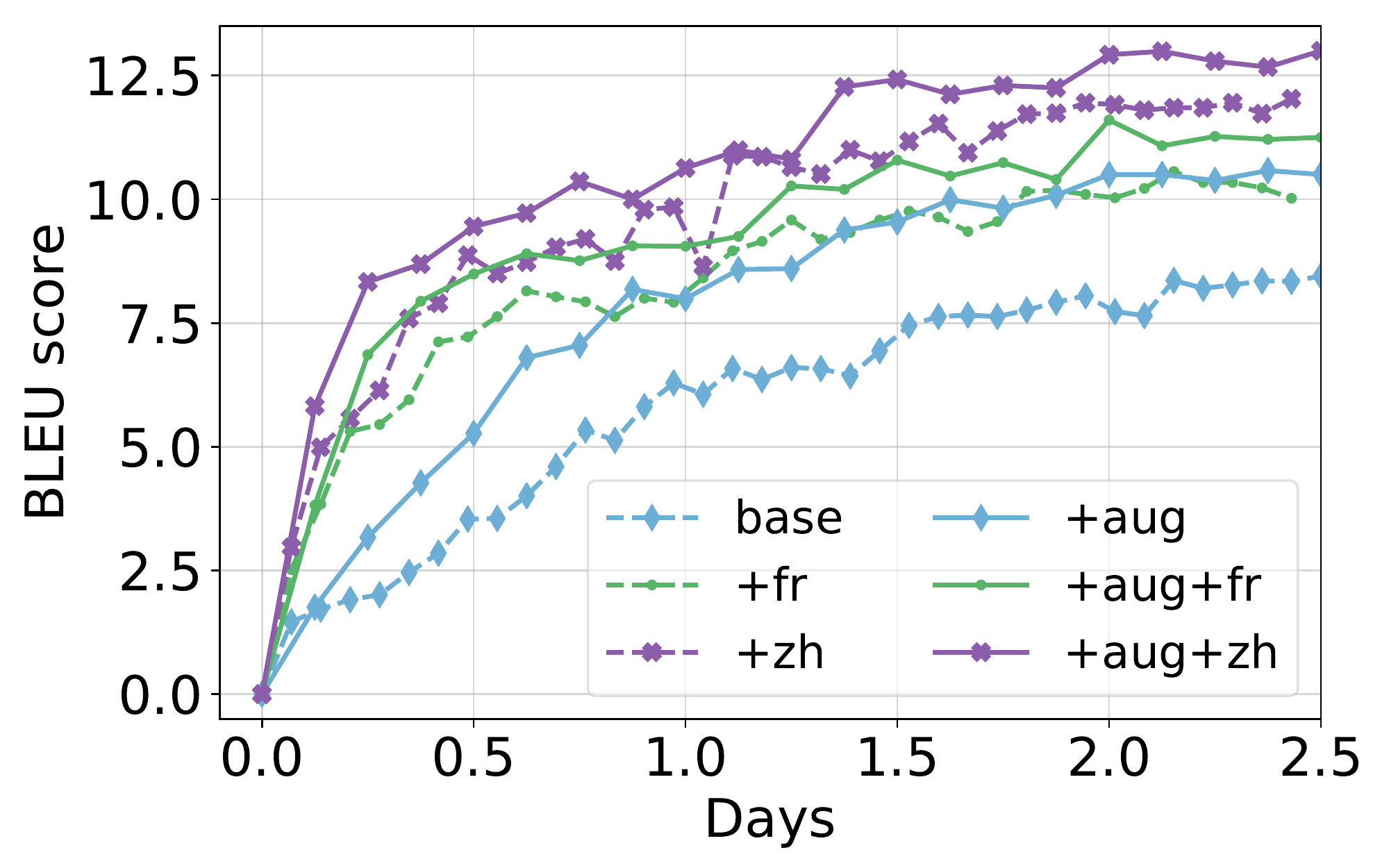}}
    \caption{The AST performance over time (without beam-search) of baseline, pretrained, and pretrained+augmented models. }
    \label{fig:time}
\end{minipage}
\end{figure}

Table \ref{tab:scores_aug} (top) shows how data augmentation affects the results of the baseline 20h AST system, as well as three of the best-performing pretrained models from Table~\ref{tab:data-results}. 
For these experiments only, we changed the learning rates of the augmented-data systems so that all models took about the same amount of time to train (see Figure \ref{fig:time}). Despite a more aggressive learning schedule, the performance of the augmented-data systems surpasses that of the baseline and pretrained models, 
even those trained on the largest ASR sets (150-hr Chinese and 300-hr English).

For comparison to other work, Table \ref{tab:scores_aug} (bottom) gives results for AST models trained on the full 160 hours of parallel data, including models with both pretraining and data augmentation. For the latter, we used the original learning schedule, but had to stop training early due to time constraints (after 15 days, compared to 8 days for complete training of the non-augmented 160h models). 
We find that both pretraining and augmentation still help, providing a combined gain of 3.8 (3.2) BLEU points over the baseline on the dev (test) set.

\begin{table}
  \begin{center}
  \begin{tabularx}{0.97\linewidth}{cl|cc|cc}
    \toprule
    & \multicolumn{1}{c}{} & \multicolumn{2}{c}{dev set} & \multicolumn{2}{c}{test set} \\
    \midrule
     & \textbf{Pretrain} & \textbf{No aug.} & \textbf{With aug.} &  \textbf{No aug.} & \textbf{With aug.}\\
     \midrule
     \multirow{4}{*}{\rotcol{20h}} 
     &-- & 10.3 & 13.0 (+2.7) & 10.2 & 13.3 (+3.1) \\
     &fr-gp & 12.5 & 13.7 (+1.2) & 12.6 & 14.3 (+1.7)\\
     &zh-ai-lrg &  14.6 & 15.5 (+0.9) & 14.3 & 15.8 (+1.5)\\
     &en-300h & 19.5 & 20.1 (+0.6) & 20.1 & 20.2 (+0.1)\\
    \midrule
    \multirow{2}{*}{\rotcol{160h}} 
     &--  & 34.1 & 36.3 (+2.2) & 34.6 & 37.3 (+2.7)\\
     &en-300h & 36.3 & 37.9 (+1.6) & 36.4 & 37.8 (+1.4)\\
  \bottomrule
  \end{tabularx}
  \end{center}
  \caption{
  BLEU scores on dev and test sets for models trained with and without data augmentation. We used either 20h of AST training data (top block) or 160h (bottom block), with various pretraining.
  }
  \label{tab:scores_aug}
\end{table}

\begin{figure}[t]
    \centering
    \includegraphics[scale=0.21]{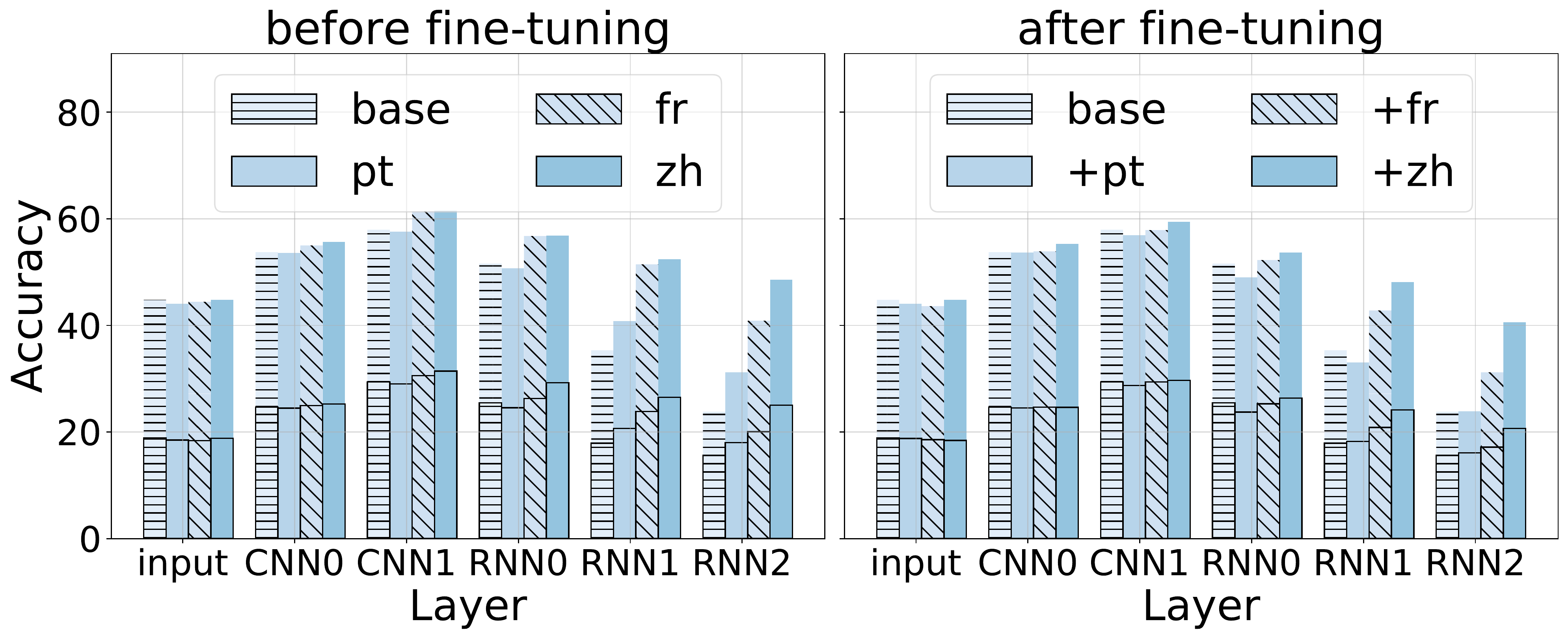}
    \caption{Phonetic classification accuracy at different layers of our ASR (left) and AST (right) models. Different color bars indicate representations extracted from models (pre)trained on different datasets (\emph{pt-gp}, \emph{fr-gp}, or \emph{zh-ai-large}). Results from the baseline AST model (without pretraining) are shown in both panels for comparison.
The bars with black edges are results on TIMIT (majority baseline: 12.9\%); the taller bars are for Spanish GlobalPhone (majority baseline: 15.2\%).}
    \label{fig:pa}
\end{figure}

\section{Analyzing the models' representations}

Finally, we hope to gain some understanding into why pretraining on ASR helps with AST, and specifically how the neural network representations change during pretraining and fine-tuning. We follow \cite{phonetic_analysis_mit} and \cite{yusheng2019_msc}, who built diagnostic classifiers \cite{hupkes2017visualisation} to examine the representation of phonetic information in end-to-end ASR and AST systems, respectively. Unlike \cite{phonetic_analysis_mit,yusheng2019_msc}, who used non-linear classifiers, we use a {\em linear} classifier to predict phone labels from the internal representations of the trained ASR or AST model.

Using a linear classifier allows us to make more precise claims: if the classifier performs better using the representation from a particular layer, we can say that layer represents the phonetic information in a more linearly separable way. Using a nonlinear classifier raises questions about how to choose the complexity of the classifier itself, and therefore makes any results difficult to interpret.

We hypothesized that pretraining allows the models to abstract away from nonlinguistic acoustic differences, and to better represent phonetic information: crucially, both in the trained language and in other languages. To test this hypothesis, we used two phone-labelled datasets distinct from all our ASR and AST datasets: the English TIMIT corpus (a language different to all of our trained models, with hand-labeled phones) and the Spanish GlobalPhone corpus (the same language as our AST source language, with phonetic forced-alignments produced using Kaldi). We randomly sampled utterances from these and passed them through the trained encoders, giving us a total of about 600k encoded frames. We used 400k of these to train logistic regression models to predict the phone labels, and tested on the remaining 200k frames.

Separate logistic regression models were trained on the representations from each layer of the encoder. Since convolutional layers have a stride of 2, the number of frames decreases at each convolutional layer. To label the frames after a convolutional layer we eliminated every other label (and corresponding frame) from the original label sequence. For example, given label sequence \emph{S\textunderscript{1} = aaaaaaann} at input layer, we get sequence \emph{S\textunderscript{2} = aaaan} at the first convolutional layer and sequence \emph{S\textunderscript{3} = aan} at the second convolutional layer and at the following recurrent layers.

Results for the two classification data sets (Figure~\ref{fig:pa}) show very similar patterns. In both the ASR and the AST models, the pretraining data seems to make little difference to phonetic encoding at the early layers, and classification accuracy peaks at the second CNN layer. However, the RNN layers show a clear trend  where phone classification accuracy drops off more slowly for models with better ASR/AST performance (i.e., zh $>$ fr $>$ pt). That is, the later RNN layers more transparently encode language-universal phonetic information.

Phone classification accuracy in the RNN layers drops for both English and Spanish after fine-tuning on the AST data. This is slightly surprising for Spanish, since the fine-tuning data (unlike the pretraining data) is actually Spanish speech. However, we hypothesize that for AST, higher layers of the encoder may be recruited more to encode semantic information needed for the translation task, and therefore lose some of the linear separability in the phonetic information. Nevertheless, we still see the same pattern where better end-to-end models have higher classification accuracy in the later layers.

\section{Conclusions}

This paper explored what factors help pretraining for low-resource AST. We performed careful comparisons to tease apart the effects of language relatedness and data size, ultimately finding that rather than either of these, the WER of the pre-trained ASR model is likely the best direct predictor of AST performance. 
Given equivalent amounts of data, we did not find multilingual pretraining to help more than monolingual pretraining,
 but we did find an added benefit from using speed perturbation to  augment the AST data. Finally, analysis of the pretrained models suggests that those models with better WER are transparently encoding more language-universal phonetic information in the later RNN layers, and this appears to help with AST.

\vfill

\pagebreak

\vfill\pagebreak

\section{Acknowledgements}
The authors wish to thank Yusheng Tian for her work on her Master's thesis at the University of Edinburgh which inspired the analysis of the change in neural network representations during pretraining and fine-tuning. Also, thanks to Dr. Yevgen Matusevych and to Ramon Sanabria for useful discussions, proof-reading and providing feedback on the paper. This work was supported in part by a James S. McDonnell Foundation Scholar Award (220020374).


\begingroup
\ninept
\bibliographystyle{IEEEtran}
\bibliography{ast}
\endgroup


\end{document}